\documentclass{article}
\usepackage{spconf,amsmath,graphicx}
\usepackage{amssymb, amsthm}
\usepackage{xcolor}
\usepackage{svg}
\usepackage{caption}
\usepackage{subcaption}
\usepackage{url}

\usepackage{cleveref}
\crefname{equation}{}{}
\Crefname{equati on}{}{}
\crefname{thm}{theorem}{theorems}
\Crefname{thm}{Theorem}{Theorems}
\crefname{clm}{claim}{claims}
\Crefname{clm}{Claim}{Claims}
\Crefname{coro}{Corollary}{Corollaries}
\Crefname{lem}{Lemma}{Lemmas}
\Crefname{sec}{Section}{Sections}
\crefname{app}{appendix}{appendices}
\Crefname{app}{Appendix}{Appendices}
\Crefname{part}{Part}{Parts}
\crefname{prop}{proposition}{propositions}
\Crefname{prop}{Proposition}{Propositions}
\Crefname{propty}{Property}{Properties}
\crefname{figure}{fig.}{figures}
\Crefname{figure}{Fig.}{Figures}
\crefname{defn}{definition}{definitions}
\Crefname{defn}{Definition}{Definitions}
\crefname{fact}{fact}{facts}
\Crefname{fact}{Fact}{Facts}
\crefname{appendix}{appendix}{appendices}
\Crefname{appendix}{Appendix}{Appendices}
\crefname{algo}{algorithm}{algorithms}
\Crefname{algo}{Algorithm}{Algorithms}
\crefname{algorithm}{algorithm}{algorithms}
\Crefname{algorithm}{Algorithm}{Algorithms}
\crefname{conj}{conjecture}{conjectures}
\Crefname{conj}{Conjecture}{Conjectures}
\crefname{obs}{observation}{observations}
\Crefname{obs}{Observation}{Observations}
\crefname{assump}{Assumption}{Assumptions}
\Crefname{assump}{Assumption}{Assumptions}
\crefname{rem}{remark}{remarks}
\Crefname{rem}{Remark}{Remarks}

\newtheorem{theorem}{Theorem}

\newtheorem{assumption}{Assumption}
\newtheorem{definition}{Definition}
\usepackage[disable]{todonotes}

\newcommand{\AG}[1]{\todo[color=red!25, inline]{ Advait: #1} \index{Advait: !#1}}

\newcommand{\GJ}[1]{\todo[color=green!25, inline]{ Gauri: #1} \index{Gauri: !#1}}

\def\x{{\mathbf x}}
\def\L{{\cal L}}
\def\w{{\mathbf w}}
\def\R{{\mathbb R}}
\def\u{{\mathbf u}}
\def\xb{{\mathbf x}}
\def\G{{\mathbf G}}
\def\O{{\mathcal O}}

\def \norm#1{\|#1\|_F}
\def \norml#1{\|#1\|_2}
\def \ip#1#2{\langle #1,#2 \rangle}
\def\E{{\mathbb E}}
\def\ceil#1{\lceil #1 \rceil}
\def \sopt{{\sqrt{\frac{2A_2}{\hat{A_1}}}}}

\def\alphab{{\bm \alpha}}

\def\mf#1{{\mathbf {#1}}}
\def\ip#1#2{\langle #1,#2 \rangle}
\def \rowmat#1#2{ \begin{bmatrix} #1 & #2 \end{bmatrix}}
\def \colmat#1#2{ \begin{bmatrix} #1\\ #2 \end{bmatrix}}
\def \fourmat#1#2#3#4{\begin{bmatrix} #1 & #2\\#3 & #4 \end{bmatrix}}

\title{Adaptive Quantization of Model Updates \\for Communication-Efficient Federated Learning}
\name{Divyansh Jhunjhunwala$^{\star}$ \qquad Advait Gadhikar$^{\star}$ \qquad Gauri Joshi$^{\star}$\qquad Yonina C. Eldar$^{\dagger}$}

\address{$^{\star}$ Carnegie Mellon University, Pittsburgh, USA, \{djhunjhu, agadhika, gaurij\}@andrew.cmu.edu \\
$^{\dagger}$ Weizmann Institute of Science, Rehovot, Israel, \{yonina.eldar@weizmann.ac.il\}}

\begin{document}

\maketitle
\begin{abstract}

Communication of model updates between client nodes and the central aggregating server is a major bottleneck in federated learning, especially in bandwidth-limited settings and high-dimensional models. Gradient quantization is an effective way of reducing the number of bits required to communicate each model update, albeit at the cost of having a higher error floor due to the higher variance of the stochastic gradients. In this work, we propose an adaptive quantization strategy called AdaQuantFL that aims to achieve communication efficiency as well as a low error floor by changing the number of quantization levels during the course of training. Experiments on training deep neural networks show that our method can converge in much fewer communicated bits as compared to fixed quantization level setups, with little or no impact on training and test accuracy.
\end{abstract}
\section{Introduction}
\label{sec:intro}

Distributed machine learning training, which was typically done in the data center setting, is rapidly transitioning to the Federated Learning (FL) setting \cite{mcmahan2017communication} \cite{kairouz2019advances}, where data is spread across a large number of mobile client devices. Due to privacy concerns, the FL clients perform on-device training and only share model updates with a central server. A major challenge in FL is the communication bottleneck due to the limited uplink bandwidth available to the clients. 

Recent work tackling this problem has taken two major directions. The first approach reduces the load on the communication channel by allowing each client to perform multiple local updates \cite{mcmahan2017communication, wang2018cooperative, wang2020overlap, haddadpour2019Local, stich2018local}, thus reducing the communication frequency between clients and server. However, this optimization may not be enough due to the large size of model updates for high dimensional models, like neural networks. The second approach deals with this problem by using compression methods to reduce the size of the model update being communicated by the clients at an update step \cite{alistarh2017qsgd, konevcny2016federated, yonina_vector_quant, wen2017terngrad, gandikota2019vqsgd, wang2018atomo, vogels2019powersgd}. However, such compression methods usually add to the error floor of the training objective as they increase the variance of the updates. Thus, one needs to carefully choose the number of quantization levels in order to strike the best error-communication trade-off.

\begin{figure}[t]
    \centering
    \includegraphics[width=0.49\textwidth]{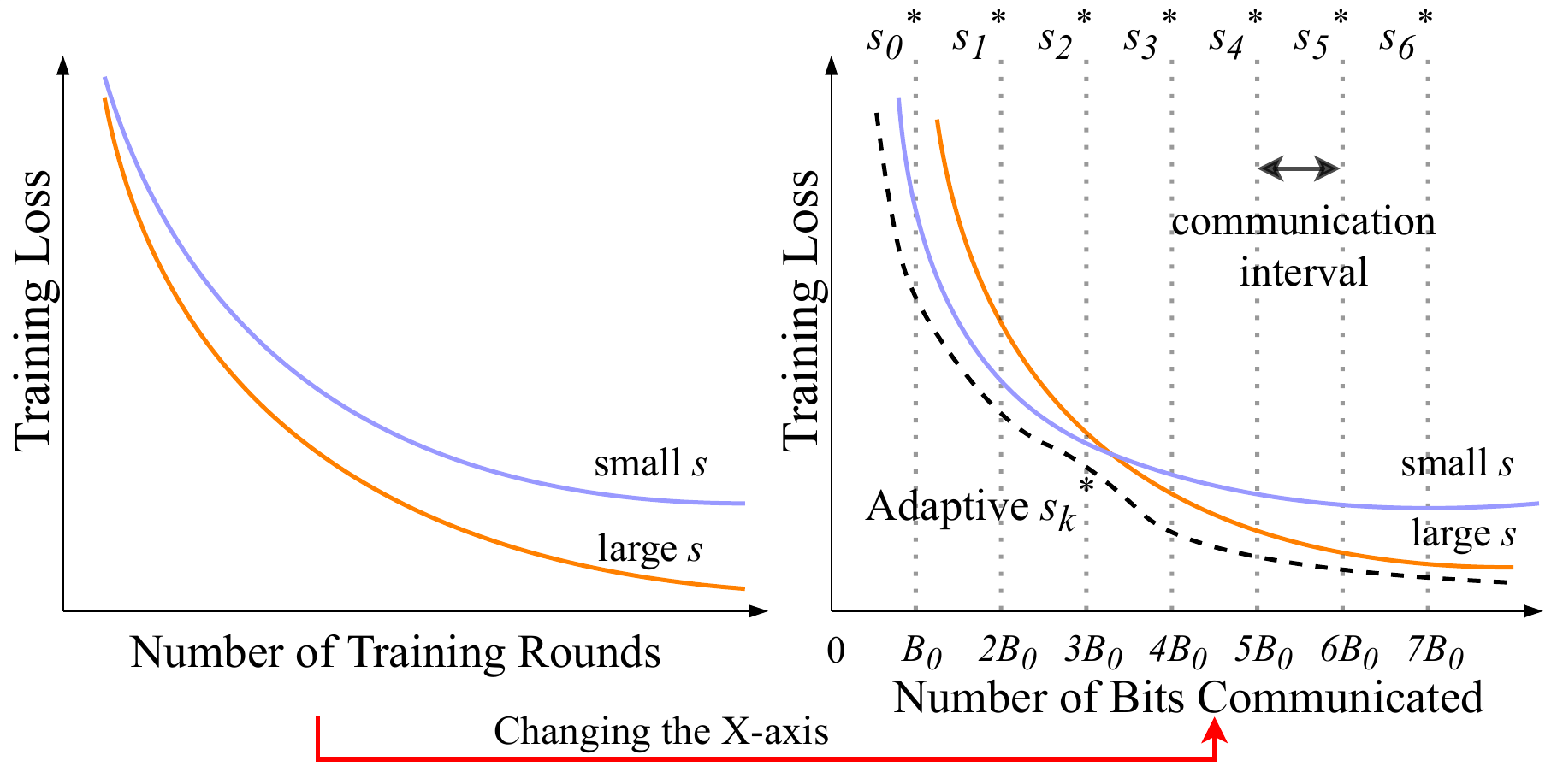}
    \caption{Viewing training in terms of bits communicated.}
    \label{Fig:schematic}
    \GJ{Write a more informative caption than just saying 'Schematic of AdaQuantFL'}
    \GJ{Increase font size}
    \GJ{Change the $x$ axis of the LHS plot to commnunication rounds. Also, the blue plot on the right hand side can be made to drop more sharply}
\end{figure}

In this work we propose AdaQuantFL, a strategy to automatically adapt the number of quantization levels used to represent a model update and achieve a low error floor as well as communication efficiency. The key idea behind our approach is that we bound the convergence of training error in terms of the number of bits communicated, unlike traditional approaches which bound error with respect to number of training rounds (see \Cref{Fig:schematic}). We use this convergence analysis to 
adapt the number of quantization levels during training based on the current training loss. Our approach can be considered orthogonal to other proposed methods of adaptive compression such as varying the spacing between quantization levels \cite{faghri2020adaptive} and reusing outdated gradients \cite{sun2019communication}.   
In \cite{wang2018adaptive}, the authors propose an adaptive method for tuning the number of local updates or the communication frequency. AdaQuantFL is a similar strategy, but for tuning the number of bits communicated per round. 
Our experiments on distributed training of deep neural networks verify that AdaQuantFL is able to achieve a given target training loss using much fewer  bits compared to fixed quantization methods. 
\section{System Model}

\label{sec:system model}

Consider a system of $n$ clients and a central aggregating server. Each client $i$ has a dataset $\mathcal{D}_i$ of size $m_i$ consisting of labeled samples $\xi_j^{(i)} = (\x_j^{(i)}, y_j^{(i)})$ for $j=1 , \dots, m_i$.
The goal is to train a common global model, represented by the parameter vector $\w \in \mathbb{R}^d$, by minimizing the following objective function:
\begin{align}
\min_{\w \in \mathbb{R}^d} \left[f(\w) = \sum_{i=1}^{n} p_i f_i(\w) = \sum_{i=1}^{n} p_i \frac{1}{m_i} \sum_{j=1}^{m_i} \ell (\w;\xi_j^{(i)}) \right], \label{eq1}
\end{align}
\noindent
where $p_i = \frac{m_i}{\sum_{i=1}^n m_i}$ is the fraction of data held at the $i$-th client and $f_i(\w)$ is the empirical risk at the $i$-th client for a possibly non-convex loss function $\ell (\w;\xi_j^{(i)})$.

\AG{Change captions make more descriptive, change the y-axis from quantization level to b* and add aline in the expts talking about reaching a loss threshold, and maybe mention 2x fewer bits. Eliminates the need for hyperparameter tuning, we do s* ourselves.}

\vspace{0.2cm} \noindent
\textbf{Quantized Local SGD.} The model is trained iteratively using the local stochastic gradient descent (local SGD) algorithm, proposed in \cite{stich2018local,wang2018cooperative}. In local SGD, the entire training process is divided into rounds consisting of $\tau$ local updates at each client. At the beginning of the $k$-th round, each client reads the current global model $\w_k$ from the central server and updates it by performing $\tau$ local SGD steps for $t=0,\cdots,\tau-1$ as follows: 
%
\begin{align}
\w_{k,t+1}^{(i)} = \w_{k,t}^{(i)} -\eta g_i(\w_{k,t}^{(i)},\xi^{(i)}),
\end{align}\label{eq2}where $\w_{k,0}^{(i)} = \w_k$ and $g_i(\w_{k,t}^{(i)},\xi^{(i)})$ 
is the stochastic gradient computed using a mini-batch $\xi^{(i)}$ sampled uniformly at random from the $i$-th client local dataset $\mathcal{D}_i$. 
%
After completing $\tau$ steps of local SGD, each client sends its update for the $k$-th round denoted by $\Delta \w_k^{(i)} = \w_{k,\tau}^{(i)} - \w_{k,0}^{(i)}$, to the central server. In order to save on bits communicated over the bandwidth-limited uplink channel, each client only sends a quantized update $Q(\Delta \w_{k}^{(i)})$, where $Q(\cdot)$ represents a stochastic quantization operator over $\mathbb{R}^d$. Once the server has received the quantized updates from all the clients, the global model is updated as follows.
\begin{align}
\w_{k+1} = \w_{k} + \sum_{i=1}^{n}p_i Q(\Delta \w_{k}^{(i)}).  \label{eq3} 
\end{align}

\vspace{0.2cm} \noindent
\textbf{Stochastic Uniform Quantizer.} In this work we consider the commonly used \cite{alistarh2017qsgd, reisizadeh2020fedpaq, suresh2017distributed} stochastic uniform quantization operator $Q_s(\w)$, which is parameterized by the number of quantization levels $s \in \mathbb{N} = \{1, 2, \dots \}$. For each dimension of a $d$-dimensional parameter vector $\w = [w_1, \dots, w_d]$, 
\begin{align}
Q_{s} (w_i) = \norml{\w}\text{sign}(w_i)\zeta_i(\w,s), \label{eq4}   
\end{align}
where $\zeta_i(\w,s)$ is a random variable given as,
\begin{align}
\zeta_i(\w,s) = 
\begin{cases}
\frac{l+1}{s} & \text{with probability } \frac{|w_i|}{\norml{\w}}s-l\\
\frac{l}{s}& \text{otherwise}.
\end{cases}
\end{align}
Here, $l \in \{0,1,2,..s-1\}$ is an integer such that $\frac{|w_i|}{\norml{\w}} \in [\frac{l}{s},\frac{l+1}{s}) $. For $\w=\mathbf{0}$, we define $Q_s(\w) = \mathbf{0}$.


Given $Q_{s}(w_i)$, we need $1$ bit to represent $\text{sign}(w_i)$ and $\lceil \log_2 (s+1) \rceil$ bits to represent $\zeta_i(\w,s)$. The scalar $\norml{\w}$ is usually represented with full precision, which we assume to be $32$ bits. Thus, the number of bits communicated by a client to the central server per round, which we denote by $C_s$, is given by
\begin{align}
\label{eq:C_s_defn}
C_s = d\lceil \log_2 (s+1) \rceil + d +32.
\end{align}
It can be shown from the work of \cite{alistarh2017qsgd,suresh2017distributed} that while $Q_s(\w)$ remains unbiased for all $s$, i.e., $\E [Q_s(\w)|\w] = \w$, the variance of $Q_s(\w)$ decreases with $s$ because of the following variance upper bound:
\begin{align}
\label{eq:grad_var}
\mathbb{E} [\|Q_s(\w)-\w\|_2^2|\w] \leq \frac{d}{s^2}\|\w\|_2^2.  
\end{align}

From \eqref{eq:C_s_defn} and \eqref{eq:grad_var}, we see that varying $s$ results in a trade-off between the total number of bits communicated $C_s$ and the variance upper bound -- $C_s$ increases with $s$ while the variance upper bound in \eqref{eq:grad_var} decreases with $s$. Building on this observation, in the next section, we analyze the effect of $s$ on the error convergence speed and use it to design a strategy to adapt $s$ during the course of training. 



\section{Trade-off Between Error and the Number of Bits Communicated}
\label{sec:conv}



%

The motivation behind adapting the number of quantization levels $s$ during training can be understood through the illustration in \Cref{Fig:schematic}. In the left plot, we see that a smaller $s$, that is, coarser quantization, results in worse convergence of training loss versus the number of training rounds. However, a smaller $s$ reduces the number of bits $C_s$ communicated per round. To account for this communication reduction, we change the x-axis to the number of bits communicated in the right plot of \Cref{Fig:schematic}. This plot reveals that smaller $s$ enables us to perform more rounds for the same number of bits communicated, leading to a faster initial drop in training loss. The intuition behind our adaptive algorithm is to start with a small $s$ and then gradually increase $s$ as training progresses to reach a lower error floor. 
To formalize this, we provide below a convergence bound on the training loss versus the number of bits communicated for any given $s$.






\vspace{0.2 cm} \noindent
\textbf{Convergence Bound in terms of Error versus Number of Bits Communicated.} 
\AG{Do we just make this Convergenve Bound as Yonina mentioned? Maybe we need to specify: convergence in terms of the number of bits communicated}
For a non-convex objective function $f(\w)$, it is common to look at the expected squared norm of the gradient of the objective function as the error metric we want to bound \cite{bottou2018optimization}. We analyze this quantity under the following standard assumptions.


\begin{assumption}
The stochastic quantization operator $Q(.)$ is unbiased and its variance is at most some positive constant $q$ times the squared $\ell_2$ norm of its argument, i.e. $\forall \text{ } \w \in \mathbb{R}^d$, $\mathbb{E} [Q(\w)|\w] = \w$ and $\mathbb{E} [\|Q(\w)-\w\|_2^2|\w] \leq q\|\w\|_2^2.$
\label{assump:1}
\end{assumption}

\begin{assumption}
The local objective functions $f_i$ are $L-$smooth, i.e. $\forall \text{ } \w, \w' \in \mathbb{R}^d$, 
$\| \nabla f_i(\w) -\nabla f_i(\w') \|_2 \leq L\|\w-\w'\|_2.$
\label{assump:2}
\end{assumption}

\begin{assumption}
The stochastic gradients computed at the clients are unbiased and their variance is bounded, that is, for all $\w \in \mathbb{R}^d$, 
$\mathbb{E} [g_i(\w,\xi^{(i)})] = \nabla f_i(\w)$ and $\mathbb{E} [\norml{g_i(\w,\xi^{(i)}) - \nabla f_i(\w)}^2 ] \leq \sigma^2.$
\label{assump:3}
\end{assumption}

\begin{assumption}
Each client $i$ has a dataset $\mathcal{D}_i$ of $m$ samples drawn independently from the same distribution (i.i.d data).
\label{assump:4}
\end{assumption}
Under these assumptions, the authors in $\cite{reisizadeh2020fedpaq}$ recently derived a convergence bound for the FL setup
described in \Cref{sec:system model}
for non-convex $\ell(\cdot;\cdot)$. 
We use this result for AdaQuantFL, however in practice our algorithm can also be successfully applied without \Cref{assump:4} (non-i.i.d data) as seen in our experiments \Cref{sec:Experimental Results}. Also
while the existing result \cite{reisizadeh2020fedpaq} studies the error convergence with respect to the number of training rounds, we bound the same error in terms of number of bits communicated, defined as follows. 
\begin{definition}[Number of Bits Communicated, $B$]
The total number of bits that have been communicated by a client to the central server until a given time instant is denoted by $B$.
\end{definition}
Since all clients participate in a training round and follow the same quantization protocol, $B$ is same for all clients at any instant. We also note that the stochastic uniform quantizer having $s$ quantization levels, satisfies \Cref{assump:1} with $q = \frac{d}{s^2}$ \cite{alistarh2017qsgd, suresh2017distributed}. Now using this definition of $B$ and our earlier definition of $C_s$ in \cref{eq:C_s_defn} we get the following theorem:
\noindent
\begin{theorem}
\label{th1}
Under Assumptions 1-4, take $Q(.)$ to be the stochastic uniform quantizer with $s$ quantization levels. If the learning rate satisfies $1 - \eta L (1+ \frac{d\tau}{s^2n}) -2\eta^2L^2\tau(\tau-1) \geq 0$, then we have the following error upper bound in terms of $B$:
\begin{align}
\frac{C_s}{B\tau}  \sum_{k=0}^{(B/C_s)-1}\sum_{t=0}^{\tau-1} \E [\norml{f(\bar{\w}_{k,t)}}^2] &\leq \nonumber\\ A_1\log_2 (4s) + \frac{A_2}{s^2} + A_3.
\label{eqn:8}
\end{align}
Here, $\bar{\w}_{k,t} = \frac{1}{n}\sum_{i=1}^{n} \w_{k,t}^{(i)}$ denotes the averaged model across all clients at each step, and 
\begin{align}
& A_1 = \frac{2(f(\mathbf{\w}_{0})-f^{*})d}{\eta B\tau}, \hspace{20pt} A_2 = \frac{\eta L d \sigma^{2}}{n}, \nonumber\\
& A_3 =  \frac{\eta^{2} \sigma^{2} (\tau-1) L^{2} (n+1)}{n} + \frac{\eta L\sigma^{2}}{n} +  A_1\frac{d+32}{d},
\end{align}
and $\w_0$ is a random point of initialization and $f^*$ is the minimum value of our objective.
\end{theorem}

The proof of Theorem 1 is deferred to Appendix A. 
This error bound allows us to see the trade-off between coarse and aggressive quantization seen in \Cref{sec:conv}, for different values of $s$. As we decrease $s$, the value of the first term in our error bound ($A_1\log_2 (4s)$) decreases but it also adds to the variance of our quantized updates which increases the second term $(A_2/s^2)$. 

\section{Proposed AdaQuantFL Strategy}
\label{sec:algorithm}
Our proposed algorithm aims at adaptively changing the number of quantization levels $s$ in the stochastic uniform quantizer such that the error upper bound in \Cref{th1} is minimized at every value $B$.
To do so, we discretize the entire training process into uniform communication intervals, where in each interval we communicate $B_0$ bits (see \Cref{Fig:schematic}). 
We now discuss how to find the optimal $s$ for each such interval.

\vspace{5pt}
\noindent
\textbf{Finding optimal $s$ for each communication interval.} We propose selecting an $s$ at any $B$ (assuming $\w_0$ as the point of initialization) by setting the derivative of our error upper bound in \cref{eqn:8} to zero. 
Doing so, we get a closed form solution of an optimal $s$ as:
\begin{align}
\label{EQN:11}
s^* = \sqrt{\frac{\eta^2 L \sigma^{2} \tau B \log_e (2)}{n(f(\mathbf{\w_0})-f^{*})}}. 
\end{align}
%
Now at the beginning of the $k$-th communication interval clients can be viewed as restarting training at a new initialization point $\w_0 = \w_k$.  
Using \cref{EQN:11} we see that the optimal $s$ for communicating the next $B_0$ bits is given by,
\begin{align}
s_k^* = \sqrt{\frac{\eta^2 L \sigma^{2} \tau B_0 \log_e (2)}{n(f(\w_{k})-f^{*})}} 
\label{eq9}
\end{align}
As $f(\w_k)$ becomes smaller the value of $s_k^*$ increases which supports our intuition that we should increase $s$ as training progresses. However, in practice, parameters such as $L$, $\sigma^2$ and $f^*$ are unknown. Hence, in order to obtain a practically usable schedule for $s_k^*$, we assume $f^*=0$ and divide $s_k^*$ by $s_0^*$ to get the approximate adaptive rule:  
\begin{align}
s_k^* \approx \sqrt{\frac{f(\w_{0})}{f(\w_{k})}} s_0^*.  \label{eq13}
\end{align}
The value of $s_0^*$ can be found via grid search (we found $s_0^* = 2$ to be a good choice in our experiments).  


\noindent
\textbf{Variable Learning Rate.} Our analysis so far assumed the existence of a fixed learning rate $\eta$. In practice, we may want to decrease the learning rate as training progresses for better convergence. By extending the above analysis, we get an adaptive schedule of $s$ for a given learning rate schedule: 
\begin{align}
\boxed{
\text{AdaQuantFL: } s_k^* \approx \sqrt{\frac{\eta_k^2f(\mathbf{\w}_{0})}{\eta_0^2f(\mathbf{\w}_{k})}} s_0^*.
\label{eq13}}
\end{align}
Here, $\eta_0$ is the initial learning rate and $\eta_k$ is the learning rate in the $k$-th interval.
%
In terms of the number of bits used to represent each element in the model update, in the $k$-th interval, AdaQuantFL uses $b_k^* =  \left \lceil \log_2 (s_k^* + 1) \right \rceil$ bits (excluding the sign bit).
\AG{corrected log to log base 2 in $b_k^*$}


\GJ{Add a sentence about convergence of AdaQuantFL}

\section{Experimental Results}
\label{sec:Experimental Results}

\begin{figure*}[h!]
    \centering
    \begin{subfigure}[b]{0.33\textwidth}\label{fig2:a}
        \centering
        \includegraphics[width=1.1\textwidth, height=\textwidth
        ]{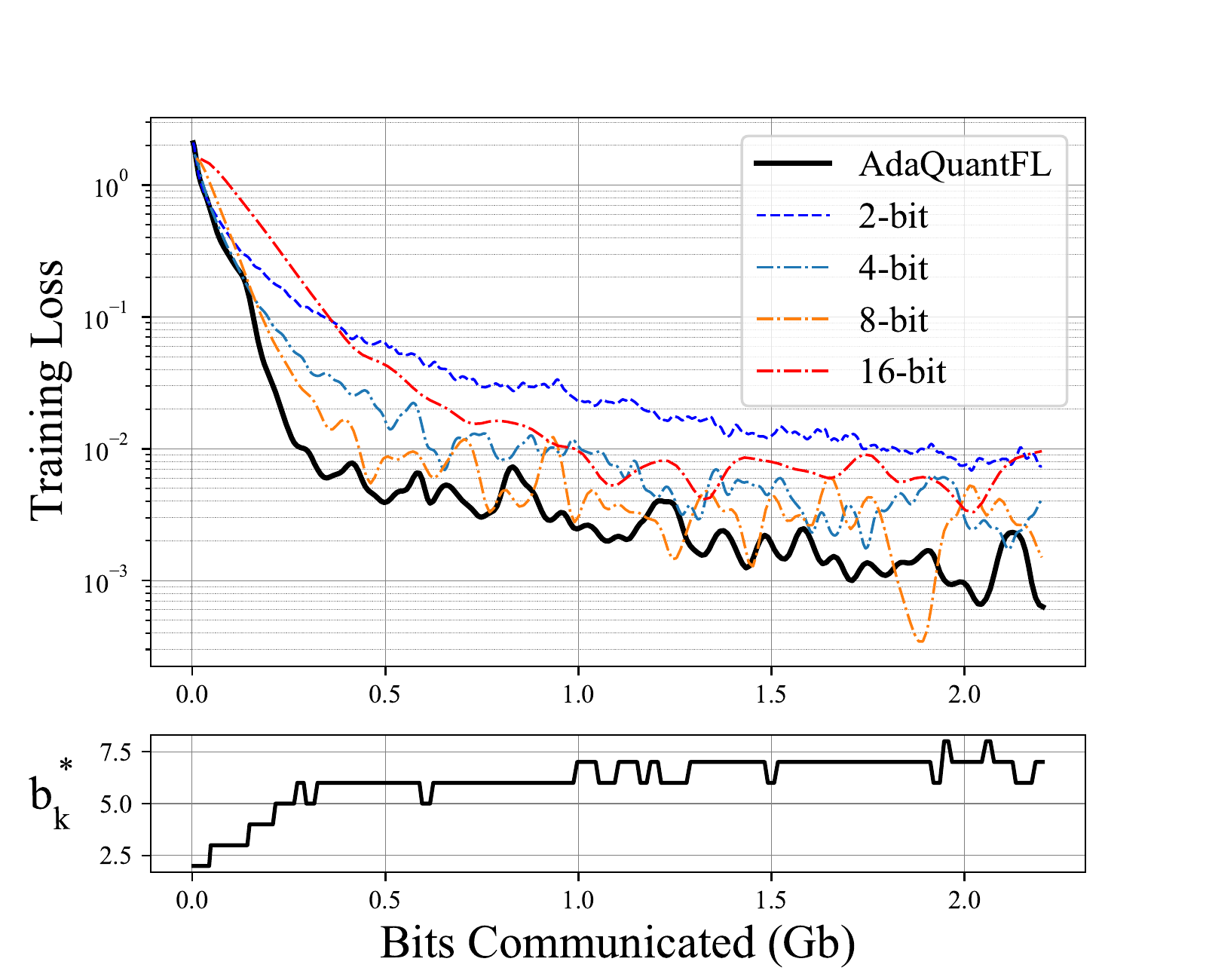}
        \caption{ResNet-18 with fixed LR, i.i.d data}
    \end{subfigure}
    \begin{subfigure}[b]{0.33\textwidth}\label{fig2:b}
        \centering
        \includegraphics[width=1.1\textwidth, height=\textwidth]{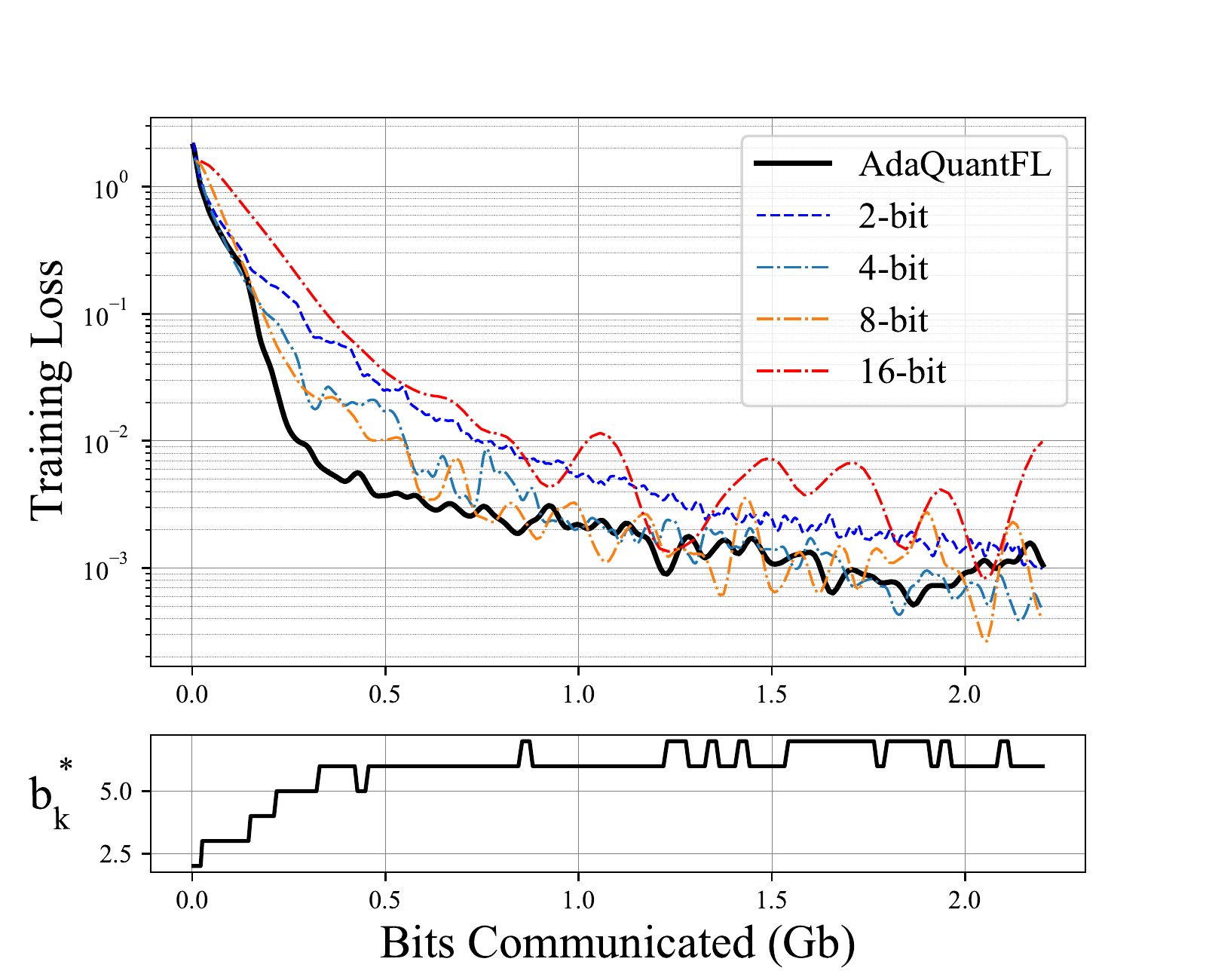}
        \caption{ResNet-18 variable LR, i.i.d data}
    \end{subfigure}
    \begin{subfigure}[b]{0.33\textwidth}\label{fig2:c}
        \centering
        \includegraphics[width=1.1\textwidth, height=\textwidth]{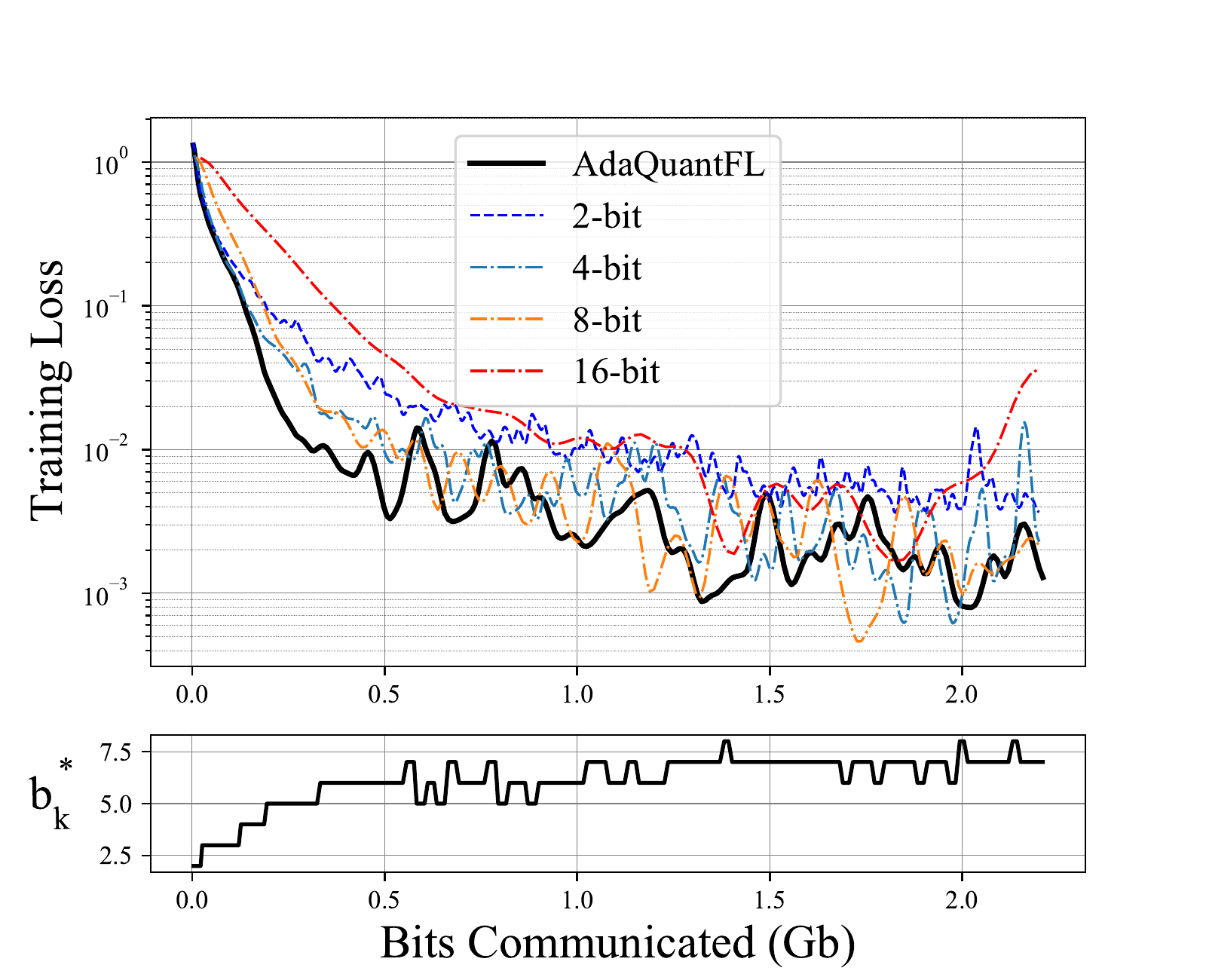}
        \caption{ResNet-18 with fixed LR, non-i.i.d data}
    \end{subfigure}
    \caption{AdaQuantFL on ResNet-18 requires a fewer bits to reach a lower loss threshold, in (a) AdaQuantFL reaches a loss of 0.02 in 0.3Gb while the 2-bit method takes 1.8Gb. Here $b_k^* = \lceil\log_2(s_k^*+1)\rceil$ (defined in \Cref{sec:algorithm}).}
    \label{fig:fig2}
\end{figure*}
\vspace{-0.3cm}

\begin{figure*}[h!]
    \newcommand{\gadhikar}{0.33\textwidth}
    \newcommand{\gadh}{0mm}
    \centering
    \begin{subfigure}[b]{\gadhikar}\label{fig1:a}
        \centering
        \includegraphics[height=\textwidth, width=1.1\textwidth]{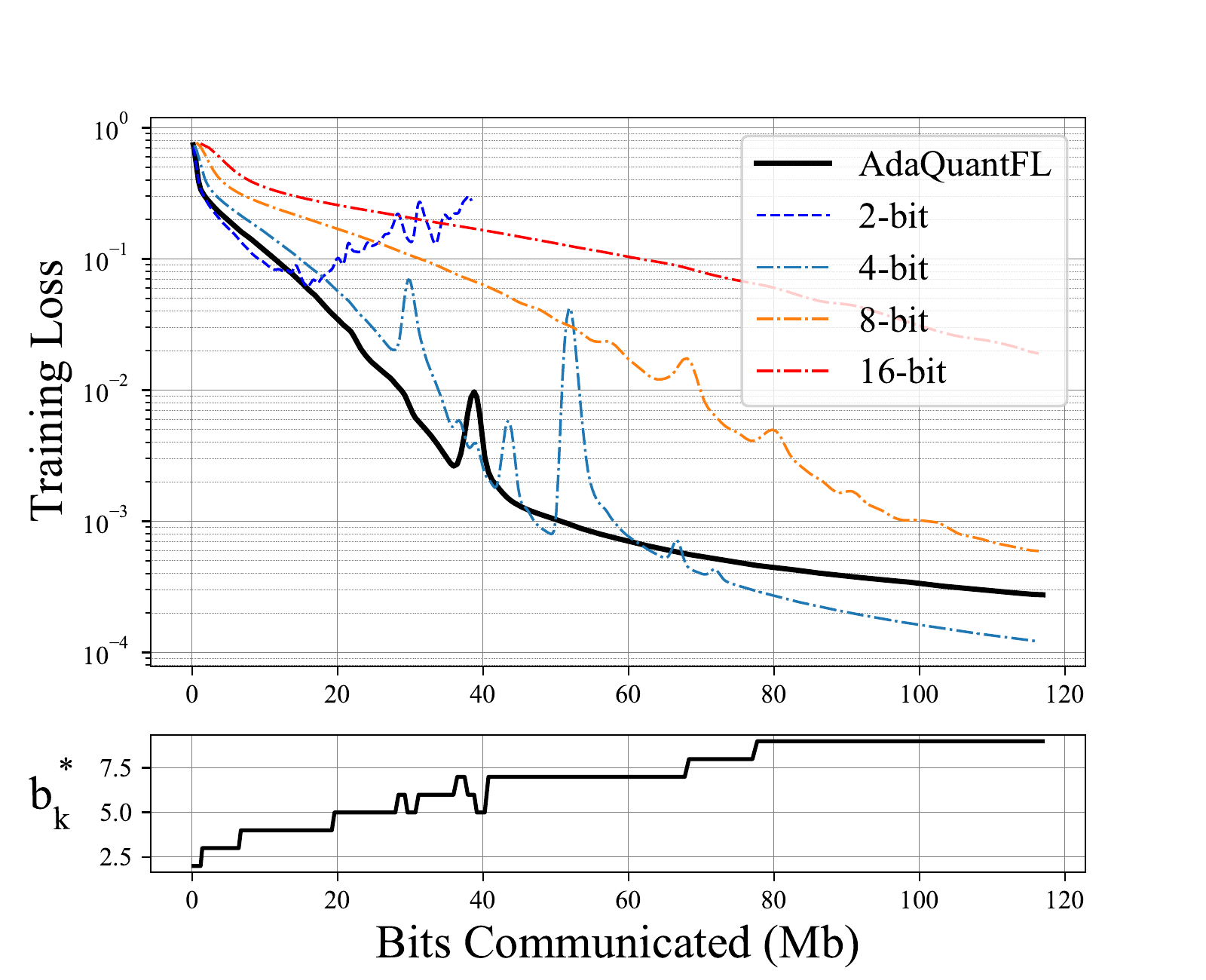}
        \caption{CNN with fixed LR, i.i.d data}
    \end{subfigure}
    \begin{subfigure}[b]{\gadhikar}\label{fig1:b}
        \centering
        \includegraphics[height=\textwidth,width=1.1\textwidth]{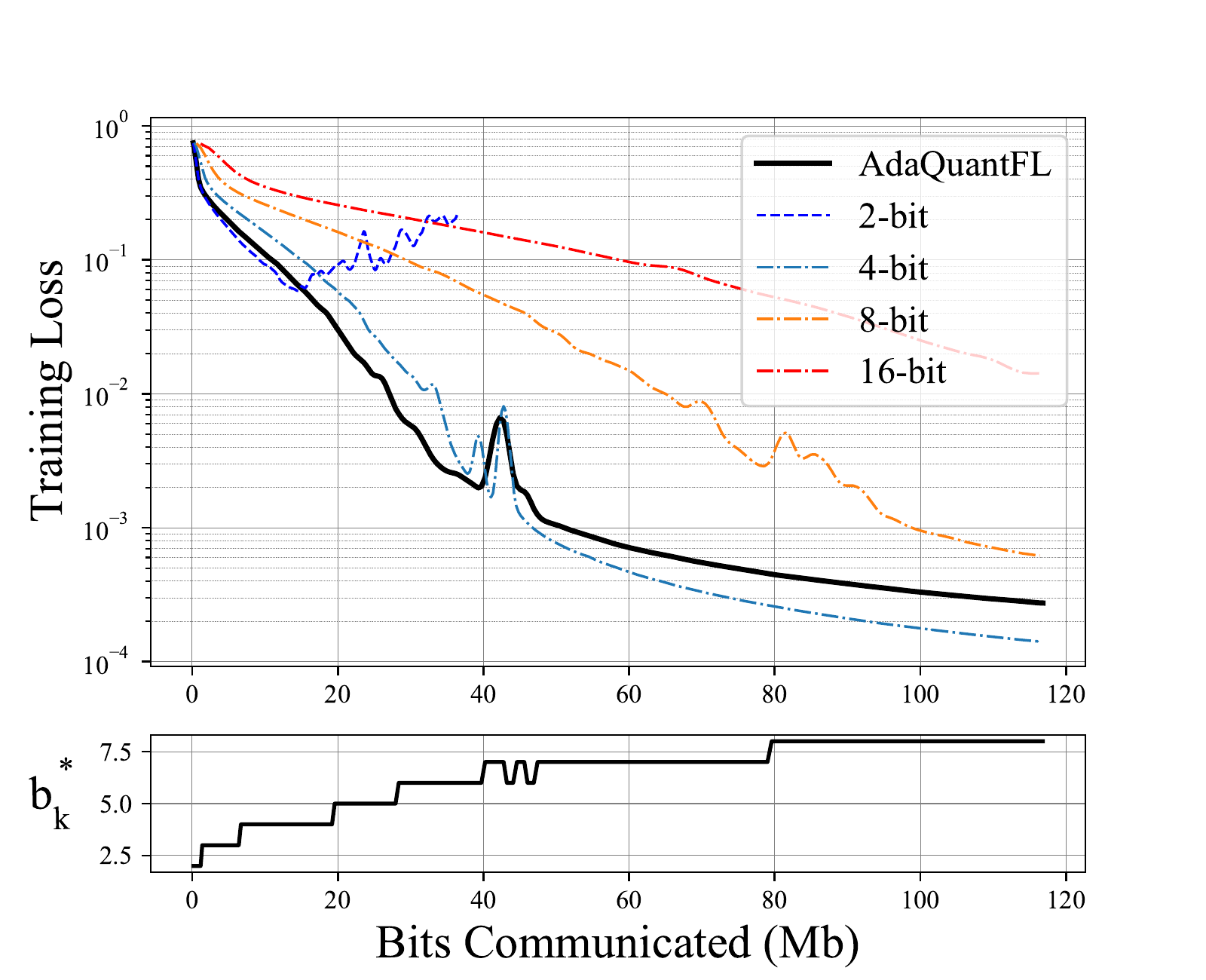}
        \caption{CNN with variable LR, i.i.d data}
    \end{subfigure}
    \begin{subfigure}[b]{\gadhikar}\label{fig1:c}
        \centering
        \includegraphics[height=\textwidth,width=1.1\textwidth]{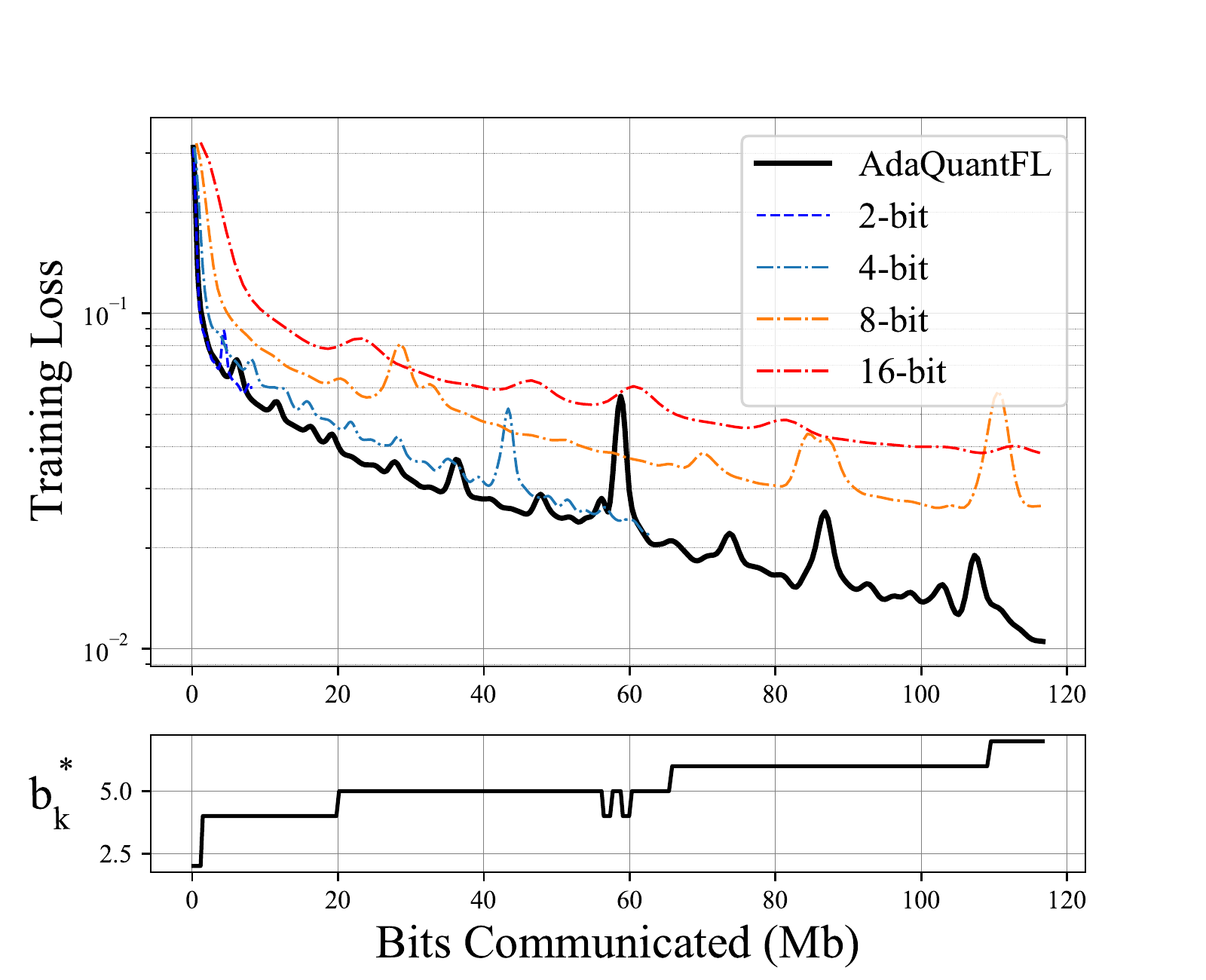}
        \caption{CNN with fixed LR, non-i.i.d data}
    \end{subfigure}
    \caption{For the Vanilla CNN, AdaQuantFL is able to achieve the lowest error floor of 0.02 for the non-i.i.d data distribution, while other methods converge at a higher error floor. Here $b_k^* = \lceil\log_2(s_k^*+1)\rceil$ (defined in \Cref{sec:algorithm}).} 
    \label{fig:fig3}
    \GJ{If space permits, I suggest adding more informative captions}
\end{figure*}

We evaluate the performance of AdaQuantFL against fixed quantization schemes using $b=\{2,4,8,16\}$ bits respectively to represent each element of the model update (excluding the sign bit) using the stochastic uniform quantizer.
The performance is measured on classification of the CIFAR-10 \cite{cifar} and Fashion MNIST \cite{xiao2017fashion} datasets using ResNet-18 \cite{7780459} and a Vanilla CNN architecture \cite{mcmahan2017communication} (referred as CNN here on) respectively.
For all our experiments we set the number of local updates to be $\tau = 10$, $\eta = 0.1$ and train our algorithm over $4$ clients for the ResNet-18 and $8$ clients for the CNN. For the variable learning rate setting, we reduce the learning rate by a factor of $0.9$ every $100$ training rounds.
We run our experiments on both i.i.d and non-i.i.d distributions of data over clients. 
%
Our experimental results verify that AdaQuantFL is able to reach an error floor using much fewer bits in most cases as seen in \Cref{fig:fig2} and \Cref{fig:fig3}. 
%
%
Additional details and figures, including test accuracy plots can be found in Appendix D.

\GJ{Mention that AdaQuantFL eliminates the need to do a brute-force search over $s$ to find the best fixed quantization level}


\section{Conclusion}
\label{sec:Conclusion}
In this paper we present AdaQuantFL, a strategy to adapt the number of quantization levels used to represent compressed model updates in federated learning. AdaQuantFL is based on a rigorous error vs bits convergence analysis. 
Our experiments show that AdaQuantFL requires fewer bits to converge during training. 
A natural extension of AdaQuantFL would be using other quantizers such as the stochastic rotated quantizer \cite{suresh2017distributed} and the universal vector quantizer \cite{yonina_vector_quant}. 

\bibliographystyle{ieee}
\bibliography{bibliography}

\cleardoublepage

\def\x{{\mathbf x}}
\def\L{{\cal L}}
\def\w{{\mathbf w}}
\def\R{{\mathbb R}}
\def\u{{\mathbf u}}
\def\xb{{\mathbf x}}
\def\G{{\mathbf G}}
\def \norm#1{\|#1\|_F}
\def \norml#1{\|#1\|_2}
\def \ip#1#2{\langle #1,#2 \rangle}
\def\ceil#1{\lceil #1 \rceil}

\def\alphab{{\bm \alpha}}
\def \sopt{{\sqrt{\frac{2A_2}{\hat{A_1}}}}}

\def\mf#1{{\mathbf {#1}}}
\def\ip#1#2{\langle #1,#2 \rangle}
\def \rowmat#1#2{ \begin{bmatrix} #1 & #2 \end{bmatrix}}
\def \colmat#1#2{ \begin{bmatrix} #1\\ #2 \end{bmatrix}}
\def \fourmat#1#2#3#4{\begin{bmatrix} #1 & #2\\#3 & #4 \end{bmatrix}}


\onecolumn
\setlength{\parindent}{0in}

\begin{center}
\section*{APPENDIX}   
\end{center}

\setcounter{section}{0}
\renewcommand\thesection{\Alph{section}}

\section{Proof of Theorem 1}
\label{Appendix:A}

We first adapt the following result from \cite{reisizadeh2020fedpaq} which states under Assumptions 1-4, for sufficiently small $\eta$ such that,
\begin{align*}
1 - \eta L (1+ \frac{q\tau}{n}) -2\eta^2L^2\tau(\tau-1) \geq 0
\end{align*}
we have after $K$ rounds of training,
\begin{align*}
\frac{1}{K\tau}  \sum_{k=0}^{K-1}\sum_{t=0}^{\tau-1} \E [\norml{f(\bar{\w}_{k,t)}}^2] &\leq \frac{2(f(\mathbf{\w}_{0})-f^{*})}{\eta K\tau} + \eta L (1+q) \frac{\sigma^2}{n} + \eta^2 \frac{\sigma^2}{n} (n+1)(\tau-1)L^2
\end{align*}
where $\bar{\w}_{k,t} = \frac{1}{n}\sum_{i=1}^{n} \w_{k,t}^{(i)}$ denotes the averaged model across all clients at each step. We note that the above result holds for any stochastic quantization operator $Q()$ that satisfies Assumption 1 with arbitrary $q$.

We assume that the stochastic uniform quantizer $Q_s(\w)$ satisfies Assumption 1 with $q = q_s$. Now by our definition of $B$, we can write $K = \frac{B}{C_s}$ (assuming $B$ mod $C_s$ = 0). Doing so, we get,
\begin{align*}
\frac{C_s}{B\tau}  \sum_{k=0}^{(B/C_s)-1}\sum_{t=0}^{\tau-1} \E [\norml{f(\bar{\w}_{k,t)}}^2] &\leq \frac{2(f(\mathbf{\w}_{0})-f^{*})C_s}{\eta B\tau} + \eta L (1+q_s) \frac{\sigma^2}{n} + \eta^2 \frac{\sigma^2}{n} (n+1)(\tau-1)L^2
\end{align*}
Now substituting $C_s = d\lceil \log_2 (s+1) \rceil + d +32 $  (using \cref{eq:C_s_defn}) and $q_s = \frac{d}{s^2} $ (using  \cref{eq:grad_var}) in RHS of the last inequality we get,
\begin{align*}
\frac{C_s}{B\tau}  \sum_{k=0}^{(B/C_s)-1}\sum_{t=0}^{\tau-1} \E [\norml{f(\bar{\w}_{k,t)}}^2] &\leq  A_1\ceil{\log_2(s+1)}+ \frac{A_2}{s^2} +A_3
\\
& \leq  A_1\log_2(4s)+ \frac{A_2}{s^2} +A_3
\end{align*}
where the last inequality follows from the fact that for $s\geq 1$ we have $\ceil{\log_2 (s+1)} \leq \log_2(4s)$. The constant $A_1$, $A_2$ and $A_3$ are defined as follows,
\begin{align}
& A_1 = \frac{2(f(\mathbf{\w}_{0})-f^{*})d}{\eta B\tau}, \hspace{20pt} A_2 = \frac{\eta L d \sigma^{2}}{n}, \nonumber\\
& A_3 =  \frac{\eta^{2} \sigma^{2} (\tau-1) L^{2} (n+1)}{n} + \frac{\eta L\sigma^{2}}{n} +  A_1\frac{d+32}{d},
\end{align}

This completes the proof for Theorem 1.

\section{Proof of Eqn. \cref{eqn:11}}

Let $F(s)$ be the objective which we want to minimize. We have,
$$F(s) =  A_1\log_2(4s)+ \frac{A_2}{s^2} +A_3$$
Now taking the first derivative we have,
\begin{align*}
\nabla F(s) =  \frac{\hat{A_1}}{s} - \frac{2A_2}{s^3}  
\end{align*}
where $\hat{A_1} = A_1 \log_2(e)$. 

Upon setting $\nabla F(s) = 0$ we get $s=\sopt$ as one of the solutions. We see that for $ s \in (0, \sopt)$, $F(s)$ is decreasing as $\nabla F(s) < 0$ and for $s \in (\sopt,\infty)$, $F(s)$ is increasing as $\nabla F(s) > 0$. This implies we get a global minima of $F(s)$ at $s = \sopt$. Substituting back the values of $\hat{A_1}$ and $A_2$ we get,
$$s^* = \sqrt{\frac{\eta^2 L \sigma^{2} B\tau \log_e (2)}{n(f(\w_{0})-f^{*})}}$$

\section{Convergence Guarantee for AdaQuantFL}
\label{sec:Convergence Guarantee}
We now provide a convergence guarantee for AdaQuantFL. In order to do so, we first state the following theorem.

\begin{theorem}[Adaptive Quantization and Variable Learning Rate Error Bound]
Assuming $K$ to be the total number of training rounds and $\eta_k$, $s_k$ to be the values of the learning rate and the quantization level in the $k$-th training round respectively, if the following condition is satisfied,
$$ \forall \text{ } k \in \{0,\cdots,K-1\}: 1- \eta_k L \left(1+ \frac{d\tau}{ns_k^2}\right) -2\eta^2L^2\tau(\tau-1) \geq 0$$
we have under Assumptions 1-4,
\begin{align}
\E & \left[ \frac{\sum_{k=0}^{K-1} \eta_k \sum_{t=0}^{\tau-1} \norml{f(\overline{\w}_{k,t)}}^2}{\sum_{k=0}^{K-1}\eta_k} \right] \nonumber\\
&\hspace{10pt}\leq \O\left(\frac{1}{\sum_{k=0}^{K-1}\eta_k}\right)
+\O\left(\frac{\sum_{k=0}^{K-1}\eta_k^2}{\sum_{k=0}^{K-1}\eta_k^2}\right)
+\O\left(\frac{\sum_{k=0}^{K-1}\eta_k^3}{\sum_{k=0}^{K-1}\eta_k}\right)
+ \O\left(\frac{\sum_{k=0}^{K-1} \eta_k^2 (d/s_k^2)}{\sum_{k=0}^{K-1}\eta_k}\right).
\end{align} \label{eqn:15}
\end{theorem}
\noindent

\textbf{Proof:}

We note here that the subscript $k$ refers to the index of the communication round, in contrast to Section \ref{sec:algorithm} where it referred to the index of the communication interval.

We also note that for the $k$-th training round the stochastic uniform quantizer with $s_k$ levels satisfies Assumption 1 with $q=\frac{d}{s_k^2}$. We now use the following result from \cite{reisizadeh2020fedpaq} (modified for the stochastic uniform quantizer) which states that under Assumptions 1-4, for the $k$-th training round if we have,
\begin{align}
1- \eta_k L (1+ \frac{d\tau}{s_k^2n}) -2\eta^2L^2\tau(\tau-1) \geq 0 \label{eqn:16}   
\end{align}
then,
\begin{equation}
 \E [f(\w_{k+1})] \leq \E [f(\w_k)] -\frac{1}{2}\eta_k \sum_{t=0}^{\tau-1} \E [\norml{\nabla f(\overline{\w}_{k,t})}^2] +\frac{\eta_k^2 L \tau \sigma^2}{2n} + \frac{\eta_k^2 (d/s_k^2) L \tau \sigma^2}{2n} +  \frac{\eta_k^3 \sigma^2 (n+1) \tau (\tau-1) L^2}{2n}   
\end{equation}
We now assume \cref{eqn:16} holds for all $k \in \{0,\cdots,K-1\}$.
Summing over all rounds $k \in \{0,\cdots, K-1 \}$ and after minor rearranging of terms we get,
\begin{equation}
\E \left[\frac{1}{2}\sum_{k=0}^{K-1}\eta_k \sum_{t=0}^{\tau-1} \norml{\nabla f(\overline{\w}_{k,t})}^2 \right]\leq f(\w_0) - f^* +\frac{L \tau \sigma^2 \sum_{k=0}^{K-1}\eta_k^2}{2n} + \frac{L \tau \sigma^2 \sum_{k=0}^{K-1}\eta_k^2 (d/s_k^2) }{2n} +  \frac{\sigma^2 (n+1) \tau (\tau-1) L^2\sum_{k=0}^{K-1}\eta_k^3 }{2n}.
\end{equation}

Dividing both sides by $\dfrac{\sum_{k=0}^{K-1}\eta_k}{2}$ we have,
\begin{align}
\E & \left[ \frac{\sum_{k=0}^{K-1}\eta_k \sum_{t=0}^{\tau-1} \norml{\nabla f(\overline{\w}_{k,t})}^2}{\sum_{k=0}^{K-1}\eta_k} \right] \nonumber\\
&\leq  
\frac{2(f(\w_0) - f^*)}{\sum_{k=0}^{K-1}\eta_k} 
+
\frac{L \tau \sigma^2 \sum_{k=0}^{K-1}\eta_k^2}{n\sum_{k=0}^{K-1}\eta_k} 
+ 
\frac{\sigma^2 (n+1) \tau (\tau-1) L^2\sum_{k=0}^{K-1}\eta_k^3 }{n\sum_{k=0}^{K-1}\eta_k}
+ 
\frac{L \tau \sigma^2 \sum_{k=0}^{K-1}\eta_k^2 (d/s_k^2) }{n\sum_{k=0}^{K-1}\eta_k}\\
& = \O\left(\frac{1}{\sum_{k=0}^{K-1}\eta_k}\right)
+\O\left(\frac{\sum_{k=0}^{K-1}\eta_k^2}{\sum_{k=0}^{K-1}\eta_k^2}\right)
+\O\left(\frac{\sum_{k=0}^{K-1}\eta_k^3}{\sum_{k=0}^{K-1}\eta_k}\right)
+ \O\left(\frac{\sum_{k=0}^{K-1} \eta_k^2 (d/s_k^2)}{\sum_{k=0}^{K-1}\eta_k}\right).
\end{align}

This completes the proof for Theorem 2.

\newpage
\subsection{Proof of Convergence:}

We assume the following conditions hold true,
\begin{align}
\lim_{K \rightarrow \infty} \sum_{k=0}^{K-1} \eta_k \rightarrow \infty, \lim_{K \rightarrow \infty} \sum_{k=0}^{K-1} \eta_k^2< \infty, \lim_{K \rightarrow \infty} \sum_{k=0}^{K-1} \eta_k^3 < \infty     
\end{align}
Now a sufficient condition for the upper bound in \cref{eqn:16} to converge to zero as $K \rightarrow \infty$ is,
\begin{align}
\lim_{K \rightarrow \infty} \sum_{k=0}^{K-1} \eta_k^2 (d/s_k^2) < \infty    
\end{align}

 Since the number of quantization levels $s_k$ will be greater than or equal to 1 for any training round, we have
\begin{align}
 \lim_{K \rightarrow \infty} \sum_{k=0}^{K-1} \eta_k^2(d/s_k^2) \hspace{5pt} \leq \hspace{5pt} d \lim_{K \rightarrow \infty} \sum_{k=0}^{K-1} \eta_k^2 \hspace{5pt} < \infty.
\end{align}

This implies as $K \rightarrow \infty$ we have,
\begin{align}
\E & \left[ \frac{\sum_{k=0}^{K-1}\eta_k \sum_{t=0}^{\tau-1} \norml{\nabla f(\overline{\w}_{k,t})}^2}{\sum_{k=0}^{K-1}\eta_k} \right] \rightarrow 0
\end{align}
 
This completes the proof of convergence.

\section{Additional Results}

\begin{figure*}[h!]
    \centering
    \begin{subfigure}[b]{0.33\textwidth}\label{2:a}
        \centering
        \includegraphics[width=1.1\textwidth, height=\textwidth
        ]{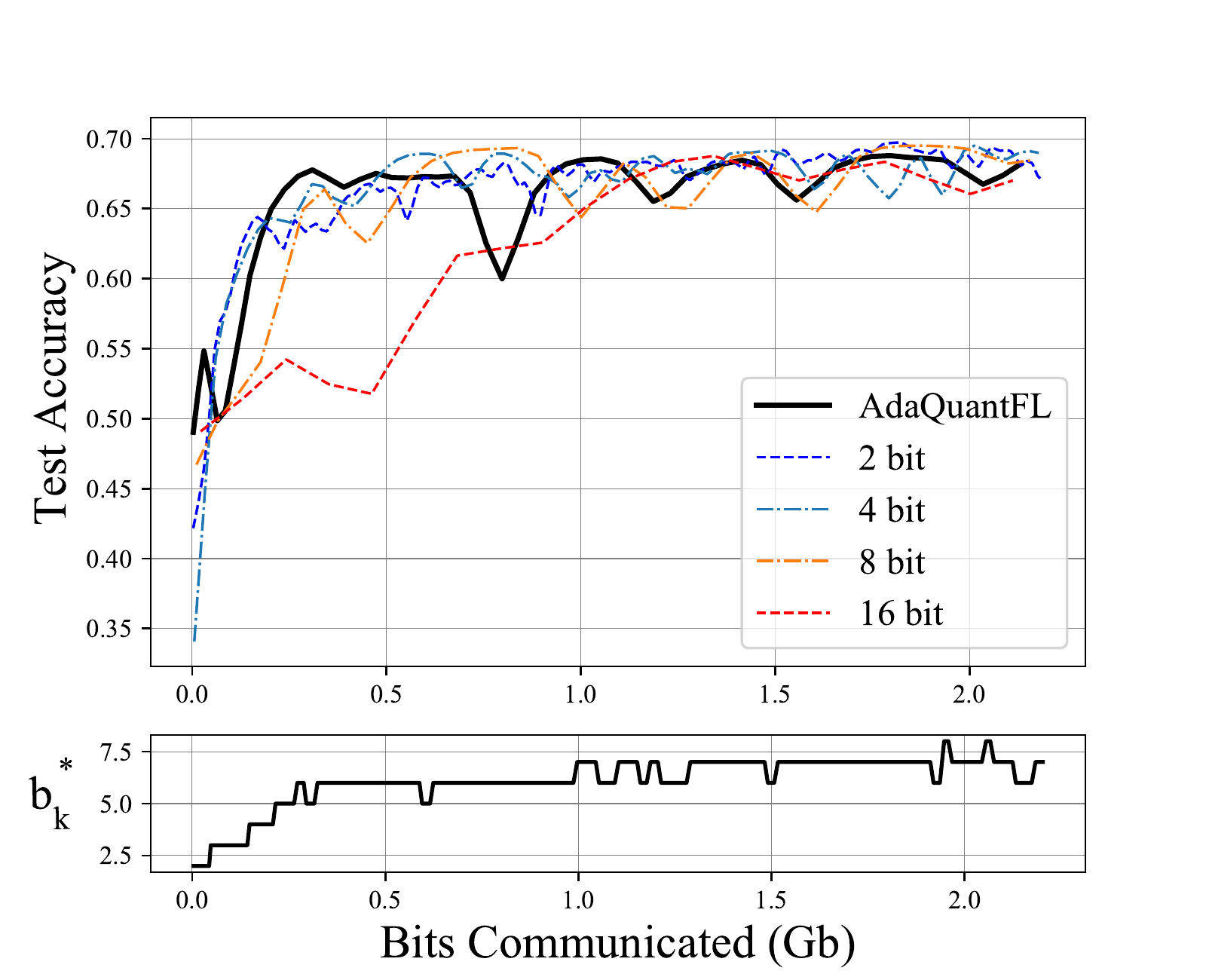}
        \caption{fixed LR, i.i.d data}
    \end{subfigure}
    \begin{subfigure}[b]{0.33\textwidth}\label{2:b}
        \centering
        \includegraphics[width=1.1\textwidth, height=\textwidth]{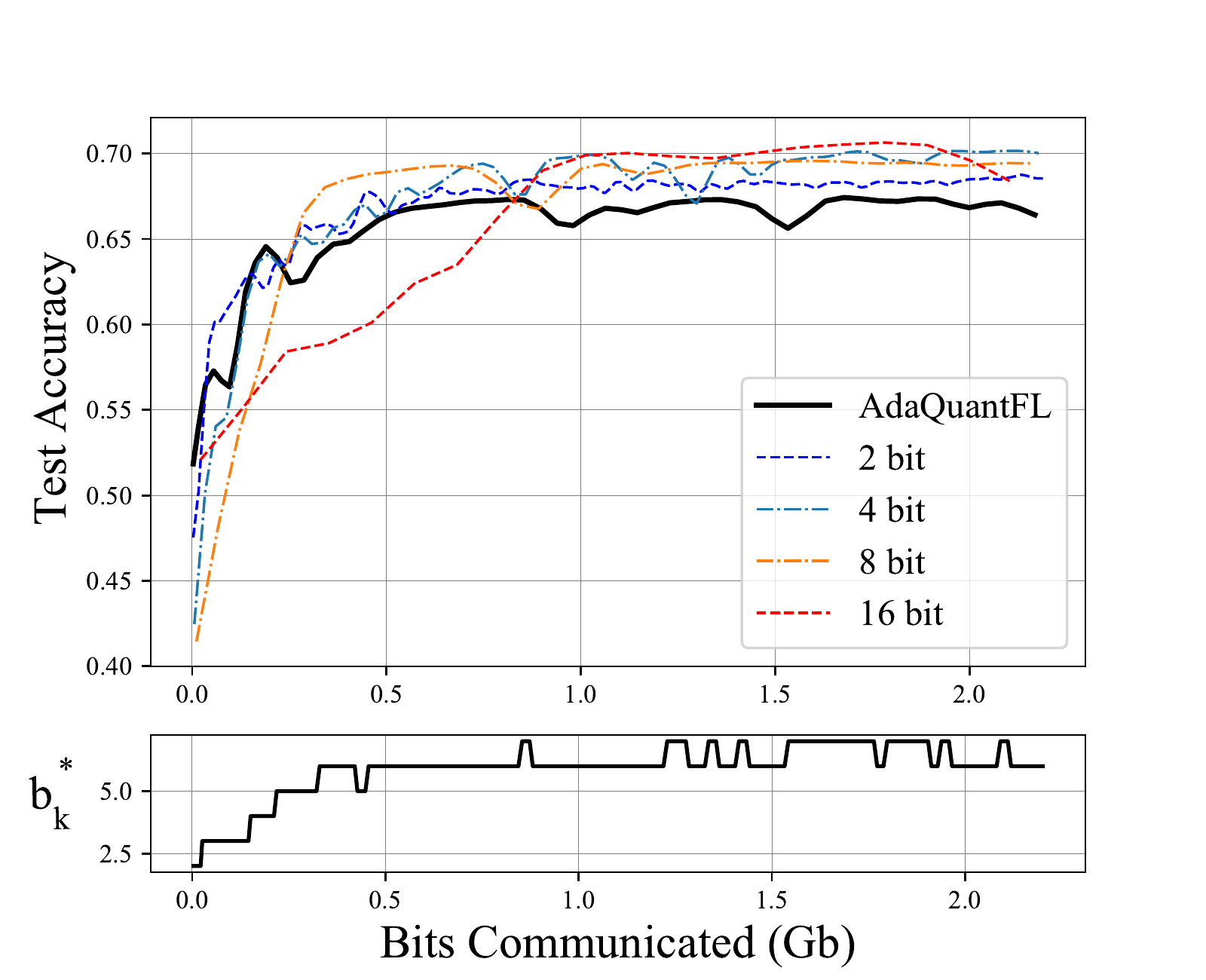}
        \caption{variable LR, i.i.d data}
    \end{subfigure}
    \begin{subfigure}[b]{0.33\textwidth}\label{2:c}
        \centering
        \includegraphics[width=1.1\textwidth, height=\textwidth]{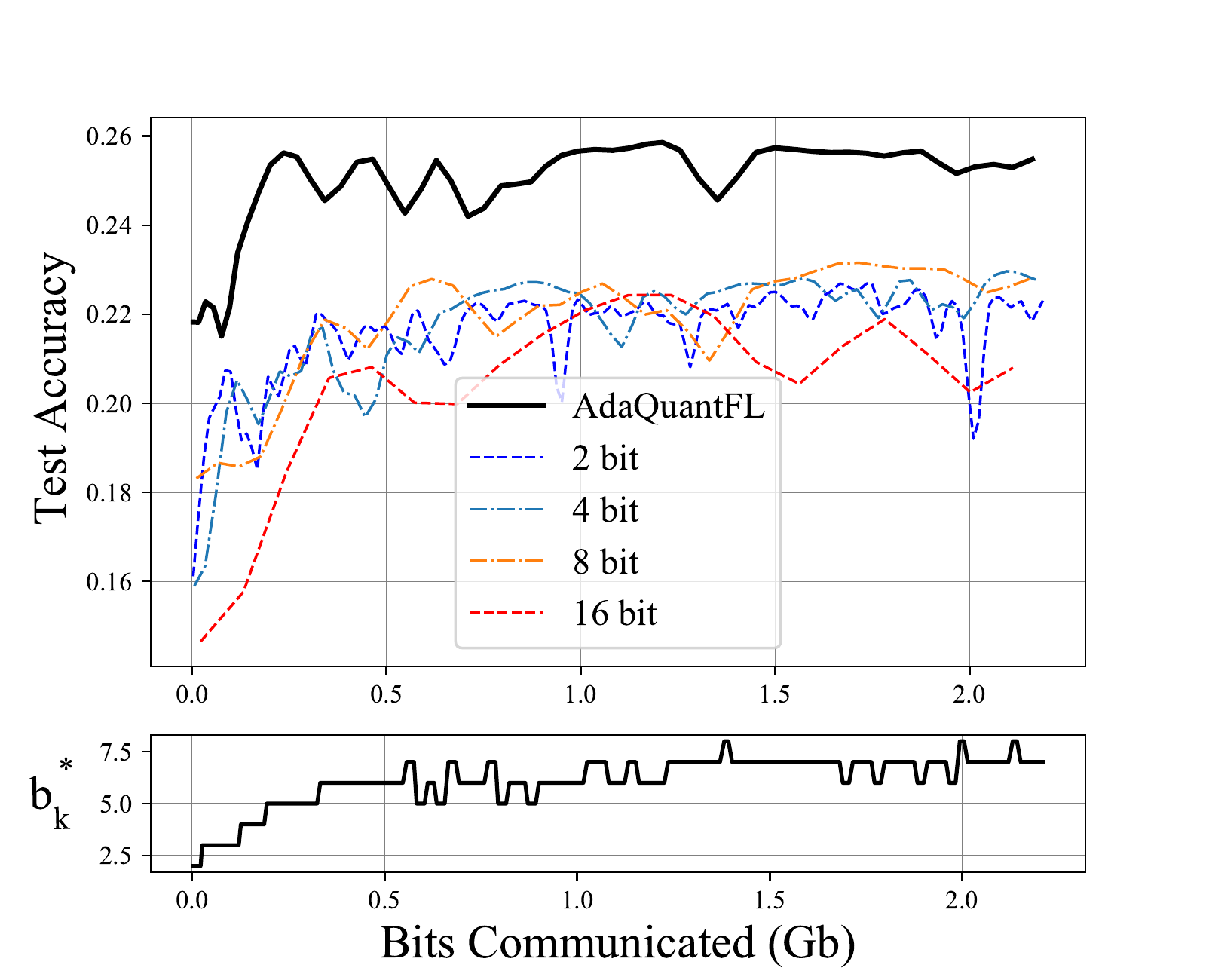}
        \caption{fixed LR, non-i.i.d data}
    \end{subfigure}
    \caption{Test Accuracy vs the number of bits communicated for  ResNet-18 on CIFAR-10}
    \label{fig:fig4}
\end{figure*}

\begin{figure*}[h!]
    \centering
    \begin{subfigure}[b]{0.33\textwidth}\label{2:a}
        \centering
        \includegraphics[width=1.1\textwidth, height=\textwidth
        ]{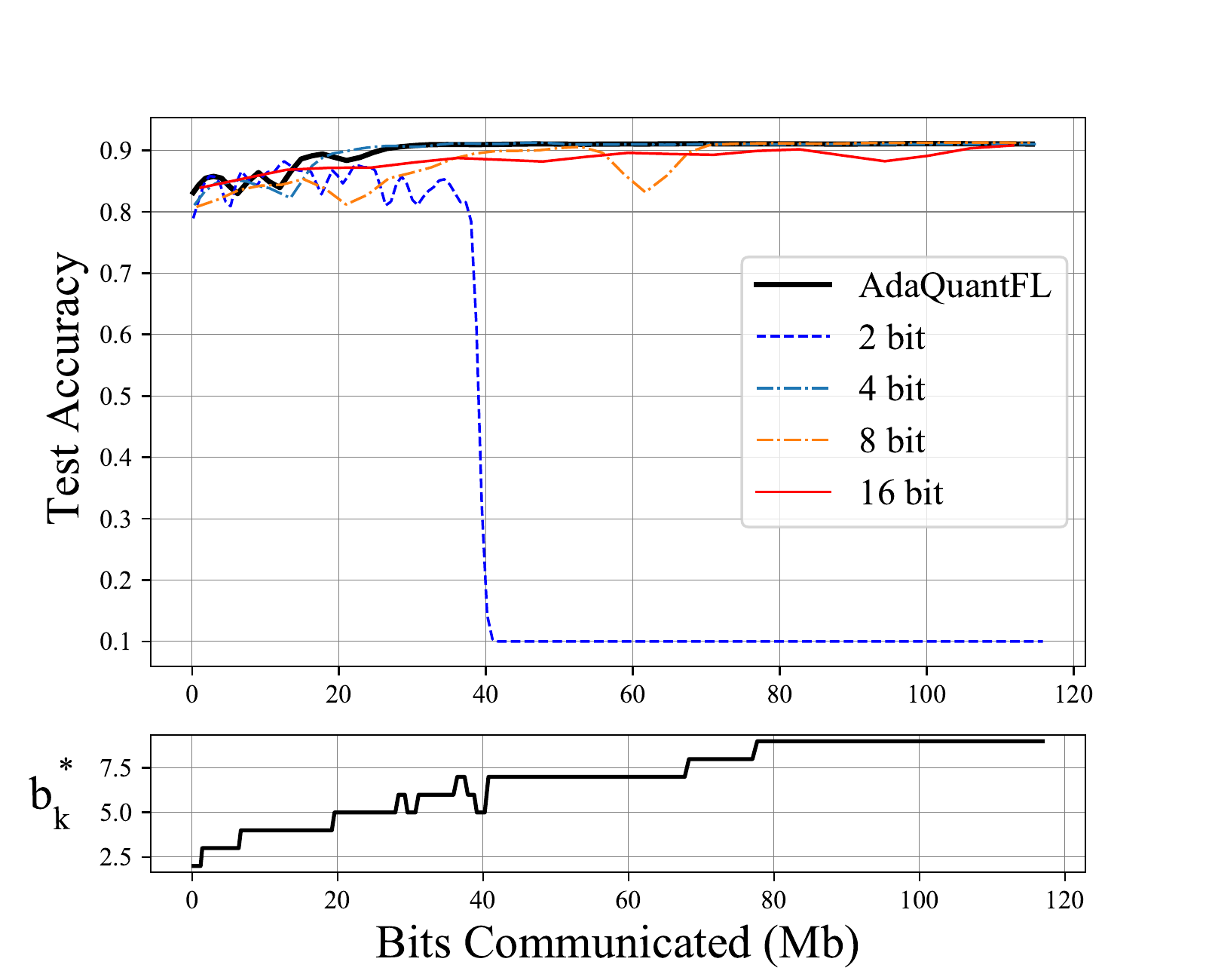}
        \caption{fixed LR, i.i.d data}
    \end{subfigure}
    \begin{subfigure}[b]{0.33\textwidth}\label{2:b}
        \centering
        \includegraphics[width=1.1\textwidth, height=\textwidth]{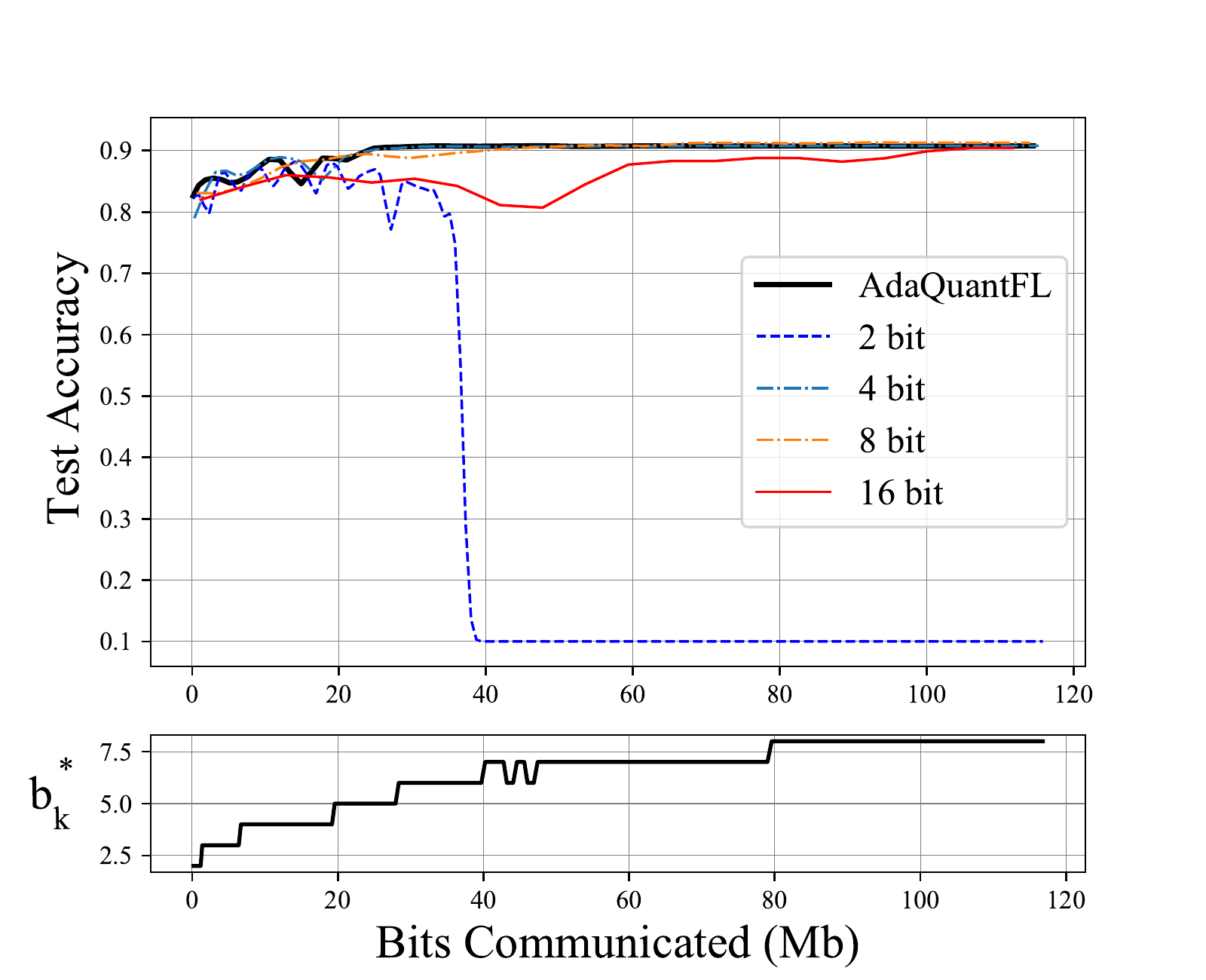}
        \caption{variable LR, i.i.d data}
    \end{subfigure}
    \begin{subfigure}[b]{0.33\textwidth}\label{2:c}
        \centering
        \includegraphics[width=1.1\textwidth, height=\textwidth]{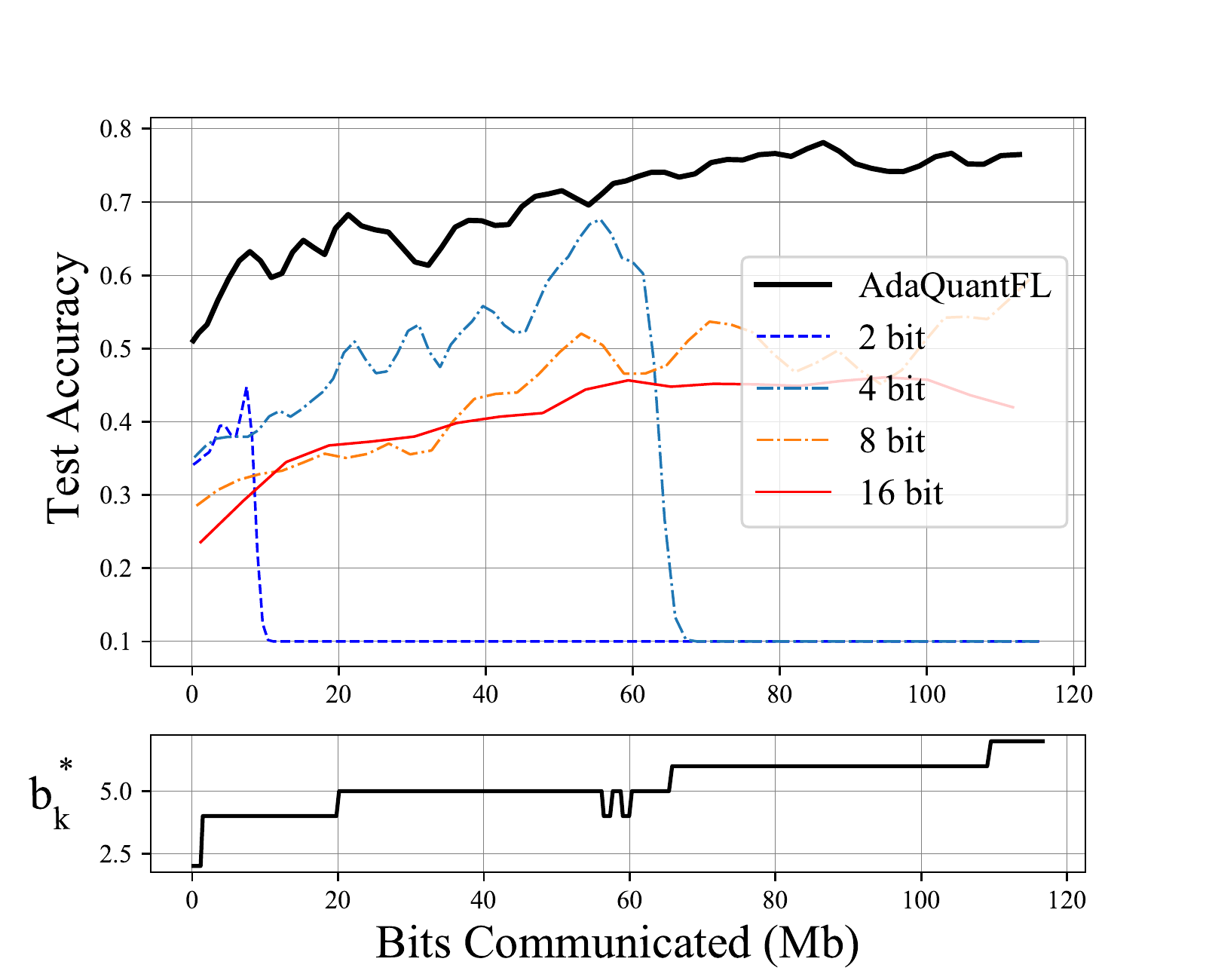}
        \caption{fixed LR, non-i.i.d data}
    \end{subfigure}
    \caption{Test Accuracy vs the number of bits communicated for Vanilla CNN}
    \label{fig:fig5}
\end{figure*}

In this section, we provide further details of our experiments and some additional results.
Figures 4 and 5 show the test accuracies for the experiments on the ResNet-18  and CNN that we trained on FMNIST and CIFAR-10 respectively.
AdaQuantFL is able to achieve a test accuracy of 69.12\% for the ResNet-18 experiment shown in \Cref{fig:fig4}(a), whereas the 16-bit quantization method achieves 69.52\%. For the CNN experiment shown in \Cref{fig:fig5}(a), AdaQuantFL reaches a test accuracy of 91.15\%, while the 16-bit method reaches 91.01\%.

For the non i.i.d settings, each dataset was sorted according to the target class labels and then partitioned equally among clients. In all experiments we fix $B_0 = 16d$ where $d$ is the dimension of our parameter vector.
The CNN architecture is inspired from \cite{mcmahan2017communication}, and consists of 2 convolutional layers with 32 and 64 channels, each followed by a max-pool and ReLU layer. The convolutional layers are followed by a linear layer of 512 with a ReLU activation and then the output softmax layer. All experiments were implemented in PyTorch \cite{pytorch} with a `gloo' distributed backend on a NVIDIA TitanX GPU.

We observe that in the case of a variable learning rate, AdaQuantFL does well for the ResNet-18 experiment shown in \Cref{fig:fig2}(b) but cannot do better than the 4-bit setting for the CNN experiment shown in \Cref{fig:fig3} (b). As observed from  \Cref{eq13}, a decreasing learning rate schedule tries to reduce $s_k^*$  while the drop in training loss does the opposite. Hence, we recommend using a conservative learning rate schedule to maximize the advantage of using AdaQuantFL.



\end{document}